# Record Counting in Historical Handwritten Documents with Convolutional Neural Networks

Samuele Capobianco
*Dipartimento di Ingegneria dell'Informazione*
*Università degli Studi di Firenze*
*Italy*
Email: samuele.capobianco@unifi.it

Simone Marinai
*Dipartimento di Ingegneria dell'Informazione*
*Università degli Studi di Firenze*
*Italy*
Email: simone.marinai@unifi.it

*Abstract*—In this paper, we investigate the use of Convolutional Neural Networks for counting the number of records in historical handwritten documents. With this work we demonstrate that training the networks only with synthetic images allows us to perform a near perfect evaluation of the number of records printed on historical documents. The experiments have been performed on a benchmark dataset composed by marriage records and outperform previous results on this dataset.

## 1. Introduction

Extracting information from historical handwritten documents coming from census, birth records, and other public or private record collections, is essential to reconstruct genealogies and to perform demographic researches [1] [2] [3] [4].

Even if a complete and automatic transcription of these documents is an obvious target for the research in handwriting document processing, some applications can require to accurately perform some sub-tasks. In particular, the automatic segmentation of records can be useful even if the actual transcription is made by trained human annotators. When dealing with large collections of documents, an accurate count of the number of records in the collection can provide valuable information to assess the amount of data available in the document images.

In document image analysis, especially when dealing with handwritten documents, the variability of the document structure can be substantial and therefore it is difficult to automatically address the document segmentation and information extraction. The record detection problem has been addressed in [5], where an EM-based layout analysis method is proposed. The approach is tested on a collection of marriage license books where each page contains a variable number of handwritten records. The records are composed by three main logical entities (body, name and tax) and the number of records in each page is variable.

Document image analysis applications have been often addressed with artificial neural networks [6] and more recently also considering CNNs [7] [8]. In this work, we address the record counting problem by means of a deep Convolutional Neural Network (CNN). We tested various network architectures, however the best results have been obtained with a modified AlexNet architecture [9], where we changed the input size and we replaced the output classification layer with one regression neuron trained to count the number of records in the page. The use of deep architectures for counting objects in images has been already proposed in the past. For instance, in [10] the task of even digits counting in synthetic images generated from the MNIST dataset is addressed with CNNs.

When training CNNs, it is essential to use a large training dataset. Clearly, annotating the record position is more complex than labeling the pages according to the number of records. However, the availability of a large training dataset annotated on the basis of the number of records in the page is not an easy assumption in real world applications.

In pattern recognition, and in particular when addressing pattern recognition tasks by learning system, common solutions to the scarcity of data are the generation of synthetic training data and the use of data augmentation. In the first case, synthetic data that simulate real patterns are generated by a suitable program. In data augmentation the number of real labeled data is increased by adding noise or distorting, in a meaningful way, the existing data. This strategy is particularly interesting when the artificial data are used for training the pattern recognition system, but the performances are computed on real data. On other hand, it is clear that data augmentation and synthetic data generation are needed when the number of available training data is limited.

In the field of document image analysis and recognition these approaches to synthetic pattern generation have been adopted, for instance, to model the process of character degradation [11] or in the field of graphical documents for generating synthetic documents for performance evaluation of symbol recognition systems [12]. In the area of handwriting recognition, cursive fonts have been used to synthetically generate handwritten documents [13].

The contribution of this paper is twofold. First, we propose a tool for the semi-automatic generation of synthetic handwritten documents containing records following a general structure. Second, we demonstrate that by training a deep architecture with artificially generated pages it is possible to estimate the number of records in real scanned

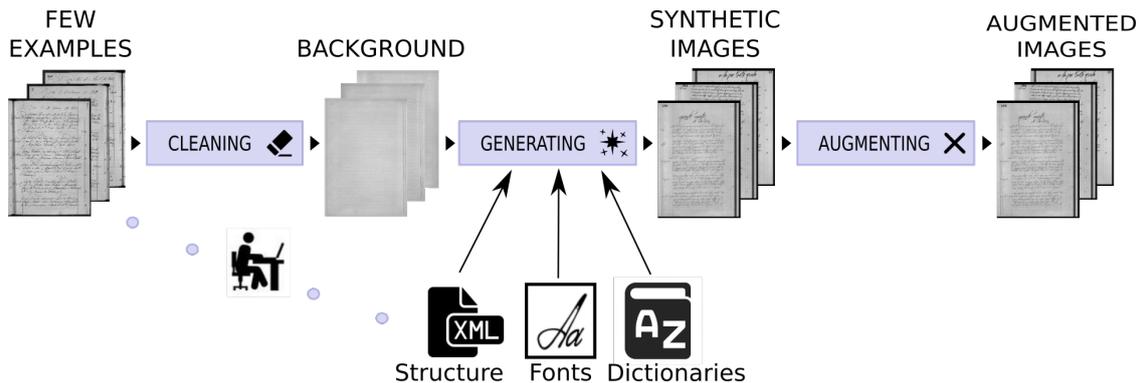

Figure 1. Main steps in the generation of synthetic handwritten documents.

historical documents with very good results.

The rest of the paper is organized as follows. In Section 2 we describe the tool developed for generating synthetic document images. In Section 3 we summarize the benchmark dataset and the experiments performed on it. Some conclusions are then drawn in Section 4.

## 2. Generation of Synthetic Data

In order to obtain a suitable number of pages to train the neural network, we implemented one flexible tool to generate synthetic pages. The aim of the tool is to generate documents that look similar to real ones both for the background and for the foreground. One important requirement of this application is that it should be flexible enough to allow the generation of a broad range of different types of documents without requiring a too complex customization, hiding to the user some details.

A general overview of the main steps involved in the generation of the synthetic data is shown in Figure 1. In general, only a few pages in the collection are required to define the record structure and to automatically extract some information from the pages. In particular, the general structure of the page is inferred by a user with a visual inspection of the documents and the expected generation rules are defined in an XML configuration file (whose information is described below).

Apart from this essential process, the other tasks are automatically performed by the tool. First, the background of some pages is extracted and subsequently used to define the substrate where the synthetic text will be printed. In the next step, a large number of synthetic pages is generated on the basis of the rules defined in the XML file, writing the text defined in the Dictionaries with some standard cursive fonts. The synthetic images are then augmented by adding some deformations that simulate the scanning process, such as page rotation and salt and pepper noise on the whole page.

The above process is quite common when generating synthetic document images. What is specific to the record counting application addressed in this paper is the generation of pages containing the desired record structure. One important point that needs to be clarified here is that for our task there is no need to model the pages at a fine grained resolution. In particular, we assume and we will verify in the experimental part, that the record counting process can be achieved regardless of the actual text printed and the actual hand writing the text. This is intuitively confirmed also by the observation that in most cases people are able to estimate the location (and therefore the number) of records in pages even without reading the text.

### 2.1. Background Extraction

The extraction of the background of actual document images in an essential pre-processing step in the generation of synthetic images. In particular, in our preliminary experiments we found that generating the pages with the algorithms described in the following sections does not provide satisfactory results when the background is significantly different (e.g. white) with respect to the collection at hand.

We therefore designed a simple module that is aimed at automatically extracting the background from actual images in the collection. To this purpose, for each page we first roughly identify the pixels belonging to the foreground by using the Otsu [14] binarization algorithm. We subsequently replace the foreground pixels with the average value of background pixels in a $20 \times 20$ window centered over each foreground pixel.

In this way, we obtain a few empty pages that can be used, like ancient palimpsests, to write new synthetic documents as described in the following section.

### 2.2. Page Generation

The XML configuration file allows the user to define the page size, the zones of the page that will contain (if existing) a page header, the structure of records and the position and number of records that can populate the page. It is also possible to define the dictionary used to write the text with the selected cursive font. To increase the noise level of the generated documents it is also possible to define the probability distribution of salt and pepper noise or other artifacts such as random white or black lines along the page.

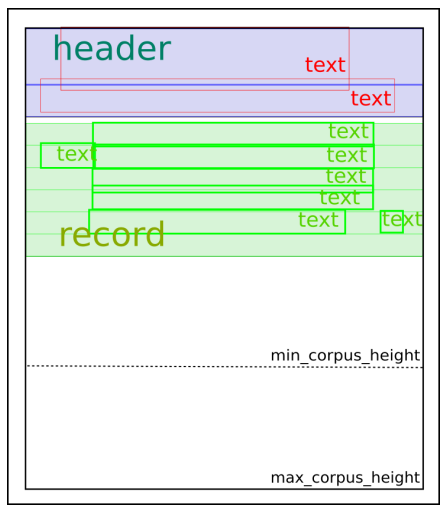

Figure 2. Page structure defined in the XML file.

In Figure 2 we graphically illustrate the main areas that are described in the configuration file. The grey area at the top of the page describes the banner of the page that often appears in documents to include, for instance, the page number or other indexing information. The record of the header, like other records, can contain mandatory fields (e.g. the page number) and other fields that are added according to a specified probability distribution.

The corpus area defines the part of the page where records are printed. Any page generated by the program will contain records in the area below the header and ending between the min_corpus_ height and the max_corpus_height delimiters. The area depicted in green defines the zone in the page where one record is generated as described in the following section.

### 2.3. Record Generation

The general idea behind the generation of records is that we focus on their visual appearance and we do not consider any possible semantics that might motivate the record organization. Therefore, rather than considering possible fields and their corresponding spatial organization, we focus on possible text lines that could belong to each record. In this way we simplify both the analysis of existing documents and the generation of synthetic records.

According to this design choice, the record structure is based on the definition of one or more types of text lines that could belong to the record. For each type of text line it is possible to define cells that can contain text (according to a given probability). Moreover, the position and size of the cells can be again fixed or changed randomly according to specified constraints.

Given the definitions of text lines one record is then defined by a variable number of text lines where it is possible to identify mandatory text lines and optional ones.

Once records have been defined, as previously summarized, the algorithm simply starts "writing" one record at a time until the min_corpus_height area is filled with records. One suitable random choice is then performed to decide whether some additional records are written in the page until, eventually, the whole area defined by max_corpus_height is filled with records. By carefully selecting the probability distributions in the model it is possible to simulate the writing process where records with different sizes are included in the page until there is enough space available. As an example, in Figure 2 we show the lines and cells of one record in the page with green boxes, while we show the lines of an header record with blue boxes.

As previously mentioned, the actual text lines are drawn by writing the text extracted from a suitable dictionary using a specific cursive font. As we will see in the experiments reported in the next section, the choice of the dictionary and of the cursive font are not critical to achieve good results. For illustrative purposes, in Figure 4 we display one page in the collection and two synthetic pages generated by the proposed tool.

## 3. Experiments

We tested the proposed approach on a benchmark dataset proposed by [5] for addressing the segmentation of historical handwritten documents. As previously discussed, one significant problem to train CNNs is the limited number of labeled samples in the dataset and in other collections available for research. We therefore generated a synthetic dataset adopting the approach described in Section 2 obtaining a synthetic training set with 81,060 pages with associated information on the number of records in each page. In particular, the training set contains pages with a number of records comprised between 3 and 9 even if the benchmark collection in [5] contains only pages with 5,6, or 7 records each. When generating the training set we used the real images only to infer the structure of the pages and the overall structure of the records as well as to extract the page background.

The initial CNN training has been performed relying only on the synthetic images while real images are used as a validation set to stop the neural network learning. This pre-trained network has been subsequently fine-tuned using some real images in the training set. Finally, a disjoint set of real images is used as a test set to compute the error rate of the system.

The model used to count the number of reacords in a page is based on the well-known Alexnet architecture [9], where we changed the input size and replaced the output classification layer with one regression neuron that is trained to count the number of records in the page. According to [10] we therefore casted the counting problem as a regression one. In the dataset used in these experiments, the input images have been rescaled to $366 \times 256$ to reduce the network input size preserving the original proportions. To generate the documents we used the "scriptina" [15], "a-glitch-in-time" [16], and "love-letter-tw" [17] cursive fonts.

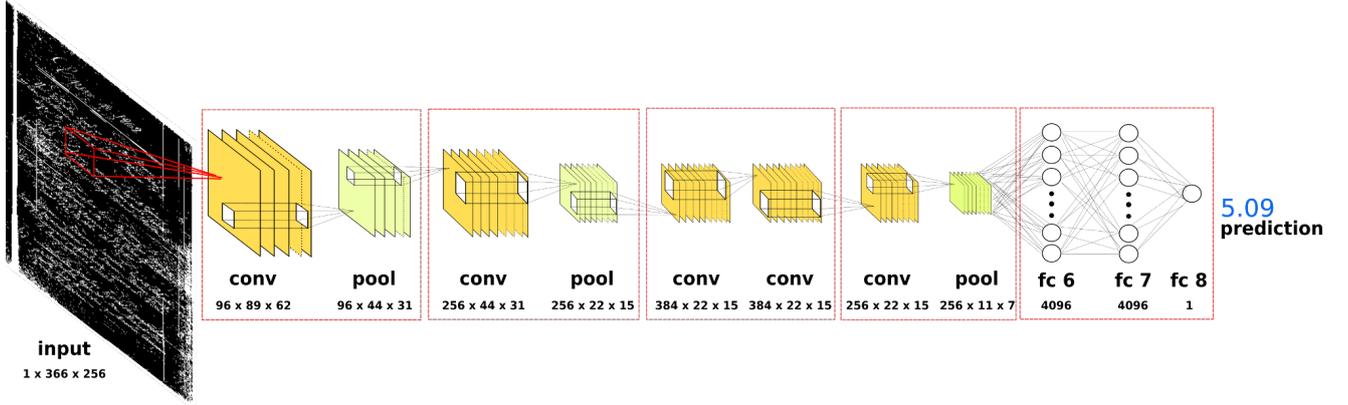

Figure 3. The proposed architecture takes a binary representation of the document as input, it is composed by five convolutional layers, three pooling layers and two fully-connected transformations to predict the number of records with one linear output neuron.

For the dictionary we considered random words from one Italian text. Therefore, the dictionary has no relationship with the textual content of the page and this is in agreement with the assumption that there is no need to read the document contents in order to infer the location of records in the page.

In Figure 3 we graphically depict the CNN architecture used in the experiments. We used binarised images as input for both training with synthetic images and testing with real ones. The image binarization has been performed with the Sauvola algorithm [18].

We performed two main experiments. In the first one, we used stratified cross validation to estimate the error rate on a larger dataset. In the second experiment we compared the results achievable by the proposed approach with those described in [5] considering the same splitting of the data in training, validation, and test datasets.

### 3.1. Five Fold Cross-validation

In this experiment, we tested the system with a five fold stratified cross validation. The dataset contains 44 pages with 5 records, 152 with 6 records and 4 with 7 records. As we can notice only few pages contain 7 records. The five folders contain, on average, 40 pages each. In each test we used 160 pages to build the validation set and perform the fine-tuning and the remaining 40 pages for evaluating the performance of the trained network on real data. In particular in the training with the synthetic images the 160 real images are used to stop the training.

In the fine-tuning, the 160 training images are augmented by generating, from each page, 8 new pages randomly rotating and adding noise obtaining a total of 1440 images. One third of these images (960) are used to fine-tune the CNN and the remaining (480) are used as validation set for the fine-tuning step to stop the CNN training.

We report in Table 1 the results obtained in each folder and as an average value in the bottom of the table. We computed three values to evaluate the system performance.

The Accuracy is the percentage of pages where the number of records is correctly identified. To provide a deeper understanding of the results we computed also the Error, that is the percentage of errors in the record count when making a decision on one page at a time. The Error is defined according to Equation (1) where $r_i$ is the actual number of records in page $i$, $p_i$ is the predicted value ($\lfloor p_i + \frac{1}{2} \rfloor$ is the rounded predicted value), and $N$ is the number of test pages.

$$Error = \frac{\sum_{i=1}^{N} \left| \lfloor p_i + \frac{1}{2} \rfloor - r_i \right|}{\sum_{i=1}^{N} r_i} \quad (1)$$

The Score value (defined in Equation (2)) is similar to the Error one but it considers the accumulated error in record counting when estimating the number of records in all the pages in the test set.

$$Score = \frac{\left| \sum_{i=1}^{N} r_i - \sum_{i=1}^{N} p_i \right|}{\sum_{i=1}^{N} r_i} \quad (2)$$

For each fold and for the average of the five folds we report the values achieved after the first training and after the fine-tuning in order to provide a feeling of the improvement obtained with the latter step.

The results achieved in the Synthetic columns are quite good if we consider that the training has been performed considering synthetic images and real data have been used only to stop the training with the validation set. As we might expect, the final fine-tuning step further improves the results.

### 3.2. Benchmark Split

In the second experiment, we compared the results obtained by our system with those presented in [5]. In order to perform a fair comparison we used the splitting of training and test data as proposed in [5]. In particular, 150 pages are used for training, 10 for validation, and 40 for test.

TABLE 1. Results with five fold cross-validation

| Folds | Accuracy | | Error | | Score | |
|---|---|---|---|---|---|---|
| | Synthetic | Fine-tuned | Synthetic | Fine-tuned | Synthetic | Fine-tuned |
| 1 | 0.927 | 1.000 | 0.013 | 0.000 | 0.015 | 0.002 |
| 2 | 0.927 | 0.976 | 0.013 | 0.004 | 0.001 | 0.003 |
| 3 | 0.825 | 0.950 | 0.030 | 0.009 | 0.025 | 0.003 |
| 4 | 0.875 | 0.925 | 0.021 | 0.013 | 0.003 | 0.013 |
| 5 | 0.789 | 1.000 | 0.036 | 0.000 | 0.030 | 0.002 |
| **Average** | **0.869** | **0.970** | **0.023** | **0.005** | **0.015** | **0.005** |

TABLE 2. Results on benchmark split

| Dataset | Accuracy | | Error | | Score | |
|---|---|---|---|---|---|---|
| | Synthetic | Fine-tuned | Synthetic | Fine-tuned | Synthetic | Fine-tuned |
| Training | 0.833 | 1.000 | 0.029 | 0.000 | 0.016 | 0.001 |
| Validation | 0.600 | 0.800 | 0.070 | 0.035 | 0.025 | 0.001 |
| **Test** | **0.900** | **1.000** | **0.017** | **0.000** | **0.016** | **0.002** |

The first training of the CNN is made as described in Section 3.1 training the network with synthetic images and using the 160 training and validation images to stop the CNN training. In the fine-tuning step we first augmented the 160 training and validation images and then used 1350 (150 × 9) pages for fine-tuninng the CNN and 90 (10 × 9) as a validation set for stopping the fine-tuning.

In Table 2 we report the results obtained on the benchmark split and in particular we show the values of the three measures previously described before and after performing the fine-tuning. To provide more information about the results obtained we report the values not only on the 40 images contained in the test set, but also on the sets of images used for training and for the validation.

Taking into account the information in Table 2 we can compare the results achieved by our system with those reported in [5]. In [5] the right number of records is predicted in the 80% of the test documents. This value corresponds to the Accuracy in Table 2 that in our case is 90% before the fine-tuning and 100% after this step. It is worth mentioning that the system described in [5] is designed to segment the records and therefore the record counting is only one information that is extracted from the segmentation. However, in this work we demonstrated that when considering the record counting task one method based on CNNs trained with synthetic data can outperform the results reported in [5].

## 4. Conclusions

In this paper we addressed the identification of the number of records in handwritten historical documents by using Convolutional Neural Networks trained on synthetic data. Thanks to the fine-tuning obtained by augmenting the images in the training set, we were able to achieve a 100% accuracy in a benchmark dataset.

We are now performing additional experiments to evaluate the robustness of the model with respect to different configurations for the document generator taking into account more fonts and various parameters settings. We are also improving the generation tool to ease the use with different datasets.

In future work we will address larger collections containing more heterogeneous documents to assess the effectiveness of the proposed approach.

## Acknowledgments

We would like to thank the authors of [5] for sharing the annotated dataset used to perform the experiments.
This work is partially supported by a research grant from Ancestry.com.

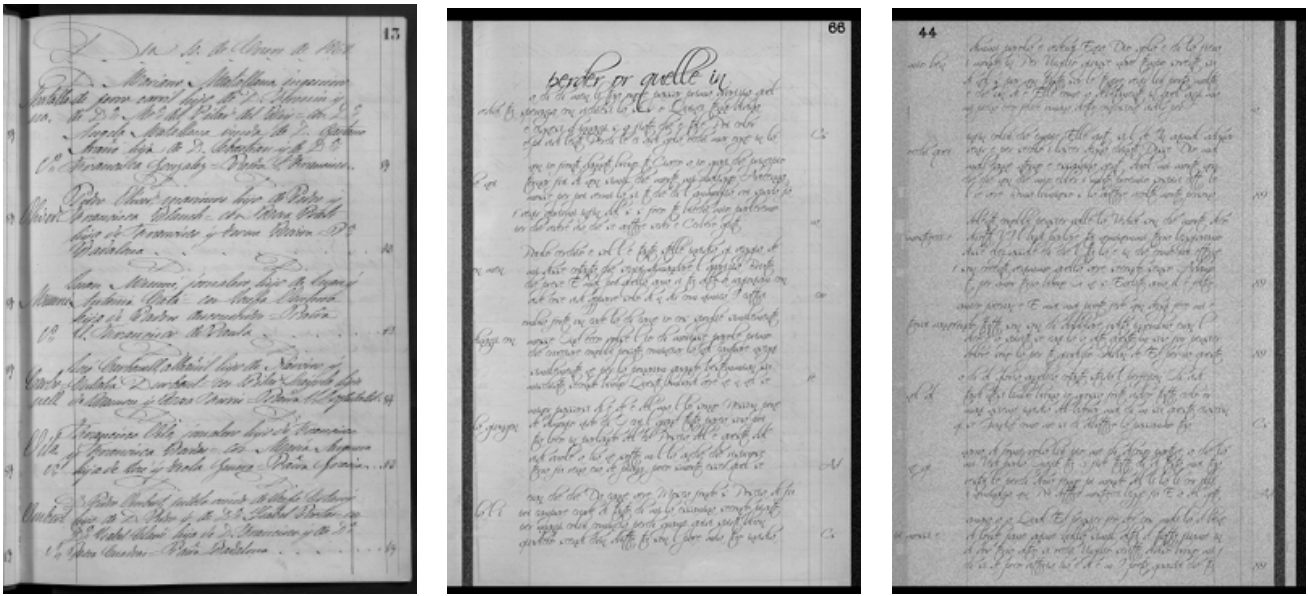

Figure 4. From left to right: real page in the benchmark collection and two generated pages.